\documentclass[wcp]{jmlr}

\jmlrproceedings{}{Preprint}

\usepackage{longtable}% for long tables
\usepackage{bm}
\usepackage[ruled]{algorithm2e}
\usepackage{multirow}

\usepackage{booktabs}
\newtheorem{def1}{Definition}[section]

\newtheorem{prop1}{Proposition}[section]
\newtheorem{exp1}{Example}[section]

\title[]{Building Rule Hierarchies for Efficient Logical Rule Learning from Knowledge Graphs}

\author{\Name{Yulong Gu} \Email{y.gu11@newcastle.ac.uk}\\
\Name{Yu Guan} \Email{yu.guan@newcastle.ac.uk}\\
\Name{Paolo Missier} \Email{paolo.missier@newcastle.ac.uk}\\
\addr Newcastle University, Newcastle upon Tyne, United Kingdom}

\begin{document}

\maketitle

\begin{abstract}
Many systems have been developed in recent years to mine logical rules from large-scale Knowledge Graphs (KGs), on the grounds that representing regularities as rules enables both the interpretable inference of new facts, and the explanation of known facts. Among these systems, the walk-based methods that generate the instantiated rules containing constants by abstracting sampled paths in KGs demonstrate strong predictive performance and expressivity. However, due to the large volume of possible rules, these systems do not scale well where computational resources are often wasted on generating and evaluating unpromising rules. In this work, we address such scalability issues by proposing new methods for pruning unpromising rules using rule hierarchies. The approach consists of two phases. Firstly, since rule hierarchies are not readily available in walk-based methods, we have built a Rule Hierarchy Framework (RHF), which leverages a collection of subsumption frameworks to build a proper rule hierarchy from a set of learned rules. And secondly, we adapt RHF to an existing rule learner where we design and implement two methods for Hierarchical Pruning (HPMs), which utilize the generated hierarchies to remove irrelevant and redundant rules. Through experiments over four public benchmark datasets, we show that the application of HPMs is effective in removing unpromising rules, which leads to significant reductions in the runtime as well as in the number of learned rules, without compromising the predictive performance.
\end{abstract}

\section{Introduction}
With the proliferation of large-scale Knowledge Graphs (KGs), such as Freebase \citep{Bollacker2008} and DBpedia \citep{Auer2007}, a large amount of systems that learn to reason in KGs for various downstream applications, including question answering \citep{Zhang2018}, fact checking \citep{Gad-Elrab2019} and knowledge graph completion \citep{Meilicke2019}, have been developed in recent years. Rule learning systems that mine logical rules from KGs are gaining popularity because the learned rules can be used to 1.) make interpretable inductive inferences, 2.) explain known facts and 3.) augment other approaches as constraints. In particular, the recent advancements of the walk-based methods \citep{Meilicke2019,Gu2020} that generate the instantiated rules containing constants from sampled paths in KGs demonstrate strong predictive performance and expressivity. However, these systems often suffer from the scalability issues where computational resources are wasted on creating and evaluating unpromising rules. In this work, we aim to address the scalability issues by applying Hierarchical Pruning Methods (HPMs) that utilize rule hierarchies to remove unpromising rules to walk-based methods.

Top-down methods \citep{Galarraga2015,Chen2016,Ho2018} generate rules by simultaneously constructing and exploring a rule hierarchy which contains the subsumption relationships between rules. By contrast, most of the walk-based methods generate rules by abstracting randomly sampled paths in KGs. Therefore, the rule hierarchy of the rules mined by walk-based methods is not readily available. As a result, the HPMs that are successfully employed in top-down methods can not be conveniently applied to walk-based methods. In this work, we propose a Rule Hierarchy Framework (RHF) that efficiently builds a proper rule hierarchy from a set of learned rules by leveraging the properties of the learned rules to simplify the process of deciding the subsumption relationships between rules. The properness of a rule hierarchy indicates that the rule hierarchy contains no redundant subsumption relationships that can be inferred via transitivity over other subsumption relationships. By adapting RHF to GPFL \citep{Gu2020} which is a walk-based system optimized in learning instantiated rules, we design and implement two HPMs that utilize the generated rule hierarchies to remove unpromising rules. Through experiments over four benchmark datasets in two settings, we demonstrate that the application of HPMs is effective in removing unpromising rules, which results in significant reductions in the runtime and the number of learned rules without compromising the predictive performance.

Our contributions are summarized as follows: 1.) we propose RHF which is the first work that aims to build proper rule hierarchies from rules mined by walk-based methods; 2.) we adapt RHF to an existing walk-based system where we design and implement two HPMs to effectively remove irrelevant and redundant rules, and 3.) we demonstrate the effectiveness of the application of HPMs through experiments over four public benchmark datasets.

\section{Related Work}
In this section, we review some of the top-down and walk-based systems, where we focus on the discussion about what and how rule pruning methods, including both flat and hierarchical ones, are implemented in existing works.

Flat Pruning Methods (FPMs) that remove a rule if the rule fails some criterion are used extensively in both top-down and walk-based methods to control the number and quality of mined rules. We here take the FPMs used in walk-based methods for example. PRA \citep{Lao2015} uses precision and coverage as criteria to filter out unqualified rules; RuDiK \citep{Ortona2018} adds a rule to the learned rule set only when the inclusion of the rule improves the overall performance of the learned rule set, and in RuleN \citep{Meilicke2018}, AnyBURL \citep{Meilicke2019} and GPFL \citep{Gu2020}, a rule is pruned if it is evaluated as unsatisfactory in terms of confidence and its coverage of positive instances.

HPMs aim to prune a rule and its descendants in a hierarchy if the rule fails some criterion. The main benefit of HPMs over FPMs is that the pruning decision on a rule can be made by merely examining the ancestors of the rule, which makes the pruning of unpromising rules even before their creation possible. By avoiding the creation and evaluation of unpromising rules, a considerable amount of computational resources are saved. We take the HPMs used in top-down methods for example. QuickFOIL \citep{Zeng2014} prevents the creation of a rule and its specializations if the rule is considered as a syntactical duplicate; AMIE+ \citep{Galarraga2015} makes an hierarchical pruning decision on a rule if the rule covers insufficient amount of positive instances; ScaLeKB \citep{Chen2016} uses a type of functional constraint to trigger the pruning, and RuLES \citep{Ho2018} uses a hybrid quality measure that takes both statistical measures and a measure computed over embeddings into consideration as the pruning criterion.
 
Because of the unavailability of rule hierarchies in walk-based methods, the more effective HPMs can not be conveniently adopted. Therefore, we need to first build a rule hierarchy from the rules mined by walk-based methods, and then implement the HPMs to achieve improved efficiency.

\section{Rule Hierarchy Framework}

In this section, we present a Rule Hierarchy Framework (RHF) that constructs a proper rule hierarchy from a set of rules, where two main problems are addressed: 1.) how to efficiently decide the subsumption relationships between rules and 2.) what subsumption relationships to include in the rule hierarchy. We achieve this by proposing a collection of subsumption frameworks and an approach to compose a proper rule hierarchy that contains no redundancies.

\begin{figure}[t]
\centering
\floatconts
{fig.1}% label for whole figure
{\caption{A small knowledge graph and its abstraction.}}% caption for whole figure
{%
\scalebox{0.8}{
\subfigure[Knowledge graph][centred]{%
\label{fig.1a}% label for this sub-figure
\includegraphics[width=0.43\textwidth]{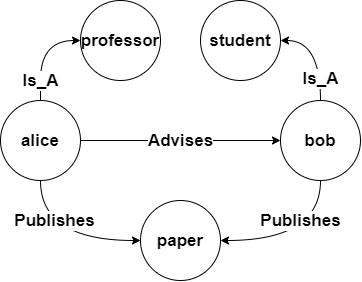}
}}\quad \quad \quad \quad% space out the images a bit
\scalebox{0.8}{
\subfigure[Abstraction][centred]{%
\label{fig.1b}% label for this sub-figure
\includegraphics[width=0.43\textwidth]{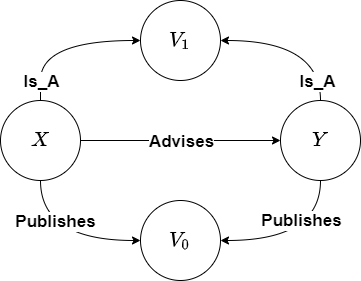}
}}
}
\end{figure}

\subsection{Preliminaries}

A knowledge graph $\mathcal{G}=(\mathcal{E}, \mathcal{R}, \mathcal{T})$ is a directed multi-graph that represents triples as relationships denoted by $r(e_i, e_j) \in \mathcal{T}$ where $r \in \mathcal{R}$ is a relationship type and $e_i, e_j \in \mathcal{E}$ are entities. As we use clausal logic for knowledge representation and reasoning, we also call a relationship $r(e_i, e_j)$ a ground atom that is composed of predicate $r$ and constants $e_i$ and $e_j$. The task of mining high-level patterns from KGs can be formulated as searching for first-order logic rules that express the regularities explaining the concepts presented in the KGs. In particular, the concepts are represented as predicates, and intuitively, the rules abstracted from the paths originated from the instances of the predicates are considered as the regularities. A path or a propositional rule, denoted by:
\begin{align*}
    & r_t(e_0, e_1), r_1(e_1, e_2), ..., r_n(e_n, e_{n+1})
\end{align*}
is a sequence of ground atoms that starts from an instance of a target predicate $r_t$. In definite Horn clause form, we can rewrite the path as:
\begin{align*}
    & r_t(e_0, e_1) \leftarrow r_1(e_1, e_2), ..., r_n(e_n, e_{n+1})
\end{align*}
which expresses a pattern where if body atoms $r_1(e_1, e_2), ..., r_n(e_n, e_{n+1})$ can be found in a KG, the existence of head atom $r_t(e_0, e_1)$ in the KG can be inferred.

\begin{exp1}
Illustrated in Figure.\ref{fig.1a} is a small knowledge graph. Consider we want to find primitive patterns for target predicate "Advises", and an instance of the predicate is relationship $Advises(alice, bob)$ which states the fact that "alice advises bob", by traversing over its neighbourhood, we can extract a set of propositional rules as the primitive patterns. For instance, an extracted propositional rule is: 
\begin{align*}
    & p_1: Advises(alice, bob) \leftarrow Publishes(alice, paper), Publishes(bob, paper)
\end{align*}
which suggests that as alice and bob have published a paper together, the statement "alice advises bob" can be inferred.
\end{exp1}

\subsection{Properties of Logical Rules}

By replacing constants with variables in a propositional rule, we can convert it into first-order logic rules. We consider a rule \textit{closed}, such as $p_1$, if all of the terms, which can be variables or constants, in the head atom also occur in the body atoms, otherwise a rule is considered \textit{open}. A rule is \textit{instantiated} if it contains at least one constant, otherwise it is an \textit{abstract} rule. Now, we define three types of first-order logic rules that are considered useful in existing works, as follows:
\begin{align*}
    & \text{\textbf{CAR}}:\quad r_t(X, Y) \leftarrow r_0(X, V_1), ..., r_n(V_n, Y)\\
    & \text{\textbf{OAR}}:\quad r_t(X, Y) \leftarrow r_0(X, V_1), ..., r_n(V_n, V_{n+1})\\
    & \text{\textbf{INSR}}:\quad r_t(X, Y) \leftarrow r_0(X, e_i), ..., r_n(V_n, e_j)
\end{align*}
where CAR and OAR stand for Closed Abstract Rules and Open Abstract Rules, respectively, and Instantiated Rules (INSRs) are open rules that contain at least one constant. For an INSR $p$, we use the deduction level, denoted by $d(p)$, to indicate the number of constants it contains. By convention, we use upper-case letters for variables and lower-case letters for constants, and symbols $X$ and $Y$ are reserved for the variables in the head atom.

\begin{exp1}
From Figure.\ref{fig.1a}, in addition to closed rule $p_1$, we can also extract open rules:
\begin{align*}
    & p_2: Advises(alice, bob) \leftarrow Is\_A(alice, professor)\\
    & p_3: Advises(alice, bob) \leftarrow Is\_A(bob, student)
\end{align*}
and by replacing all of or a part of constants with variables in rules $p_1$, $p_2$ and $p_3$, we have:
\begin{align*}
    & p_4: Advises(X, Y) \leftarrow Publishes(X, V_0), Publishes(Y, V_0)\\
    & p_5: Advises(X, Y) \leftarrow Is\_A(X, V_1) \quad p_6: Advises(X, Y) \leftarrow Is\_A(Y, student)
\end{align*}
where $p_4$ is a CAR abstracted from $p_1$; $p_5$ is an OAR converted from $p_2$, and $p_6$ is an INSR abstracted from $p_3$ and $d(p_6) = 1$.
\end{exp1}

Unlike logical rules mined using classic Inductive Logic Programming (ILP) \citep{Muggleton2012} approaches with relaxed syntactic biases, the form of the rules extracted from KGs is restricted in accordance with the ontology of the KGs and the nature of paths. In particular, we define two important properties about the logical rules generated from paths.

\begin{def1} [Connectedness]
As the paths in KGs are connected, in the sense that adjacent relationships share a  constant, the rules that are generalized from the paths are also connected, where adjacent atoms are connected via a connecting term. 
\end{def1}

The connectedness of rules is desirable in that it ensures that the body atoms of a rule are tied to each other and to the head atom via a chain of connections.

\begin{def1} [Straightness]
A rule is considered straight if for any term $t$ in the rule, $t$ occurs at most twice. 
\end{def1}

As in other works \citep{Meilicke2019,Gu2020}, the straightness of rules prevents the generation of cycles and syntactical equivalence that compromises system performance. In addition, we restrict our discussion to the KGs that only contain binary predicates. The rule space of a rule learner is a set containing all possible rules that can be produced by the learner. We denote by $\mathcal{F}$ a set of rules or a rule space. We model a rule hierarchy as a set of subsumption relationships $\Phi=\{\phi_0, ..., \phi_n\}$ where a subsumption relationship $\phi=(p, p')$ implies that $p$ subsumes $p'$. We also use notation $(\mathcal{F}, \preceq)$ to conveniently describe a hierarchy which contains the subsumption relationships resolved by a subsumption framework $\preceq$ over the rules in $\mathcal{F}$. To efficiently build a rule hierarchy from a set of rules, the subsumption frameworks with efficient proof procedure and the approach that constructs rule hierarchies with minimum redundancy are needed.

\subsection{Efficient Subsumption Framework}

In this section, we present a subsumption framework that leverages the connectedness and straightness of rules to simplify the proof procedure of a variant of the $\theta$-subsumption \citep{Maloberti2004} for improved efficiency.

$\theta$-subsumption, as a decidable approximation of logical entailment, is one of the most important subsumption frameworks employed in ILP works. A rule $p$ $\theta$-subsumes rule $p'$, denoted by $p \preceq_{\theta} p'$, iff:
\begin{align} \label{eq1}
    & \exists \theta : p \theta \subseteq p'
\end{align}
where $\theta$ is a substitution that replaces variables by terms. For instance, we have $p_5 \preceq_{\theta} p_6$ with $\theta=\{V_1 \backslash student\}$. $\theta$-subsumption is inconsistent with our assumption that all mined rules are straight in that it allows different variables to refer to the same entities. For consistency, we instead employ $\theta$-subsumption under Object Identity (OI-subsumption) \citep{Ferilli2002}, denoted by $\preceq_{OI}$. A rule $p$ OI-subsumes rule $p'$ iff $p \preceq_{\theta} p'$ and all variables in $p$ after substitution refer to different entities. 

\begin{exp1} \label{exp2.3}
Figure.\ref{fig.1b} illustrates an abstraction of the KG in Figure.\ref{fig.1a}. In Figure.\ref{fig.1b}, only CARs and OARs are explicitly demonstrated as paths while INRS are implied by replacing variables with corresponding constants. Given the target predicate "Advises", we extract rules:
\begin{align*}
    & p_7: Advises(X, Y) \leftarrow \quad p_8: Advises(X, Y) \leftarrow Publishes(X, V_0)
\end{align*}
where $p_7$ is known as the top rule that subsumes all of the rules having target predicate "Advises". Consider we have a rule set $\mathcal{F}=\{p_4, p_7, p_8\}$, rule hierarchy $(\mathcal{F}, \preceq_{OI})$ is then $\{(p_8, p_4), (p_7, p_8), (p_7, p_4)\}$.
\end{exp1}

The proof procedure of deciding whether a rule $p$ OI-subsumes rule $p'$ can be summarized as follows: $p'$ is first skolemized by having all of its variables replaced with the constants not occurring in $p$ and $p'$. We denote by $S(p')$ the skolemized $p'$ . Then an atom in $p$, denoted by $p[i]$ where $[i]$ is an element accessor that returns the i-th atom in a rule, is tested for elimination, that is if $p[i]$ subsumes an atom in $S(p')$ with a substitution that assigns variables to different entities, $p[i]$ is eliminated from $p$. This process is recursively performed until all atoms in $p$ are eliminated and then we conclude $p \preceq_{OI} p'$. If the elimination test fails, it backtracks to test $p[i]$ against another atom in $S(p')$, and if all comparable atoms in $S(p')$ fail the test, we conclude that $p \not \preceq_{OI} p'$. 

% \begin{exp1} \label{exp2.4}
% Consider we want to decide whether $p_8$ OI-subsumes $p_4$, we first skolemize $p_4$ as:
% \begin{align*}
%     & S(p_4): Advises(sk_0, sk_1) \leftarrow Publishes(sk_0, sk_2), Publishes(sk_1, sk_2)
% \end{align*}
% where substitution $\{X \backslash sk_0, Y \backslash sk_1, V_0 \backslash sk_2 \}$ is applied. By applying $\{X \backslash sk_0, Y \backslash sk_1 \}$ to $p_8$, we prove $p_8[0] \preceq_{OI} S(p_4)[0]$, thus $p_8[0]$ is eliminated and $p_8[1]$ becomes $Publishes(sk_0, V_0)$. When testing $p_8[1]$ against $S(p_4)[2]$, the testing fails as $sk_0$ can not be reduced to $sk_1$. We then backtrack to test $p_8[1]$ against $S(p_4)[1]$ where we have $p_8[1] \{V_0 \backslash sk_2 \} = S(p_4)[1]$. With $p_8[1]$ eliminated, all atoms in $p_8$ are eliminated, and hence we conclude that $p_8 \preceq_{OI} p_4$.
% \end{exp1}

% The backtracking process causes inefficiency in resolving the OI-subsumption. By leveraging the connectedness and straightness of rules, we propose a constrained version of the OI-subsumption that simplifies the proof procedure by replacing the backtracking process with a position constraint.

\begin{exp1} \label{exp3.4}
Given rules:
\begin{align*}
    & p_9: r_t(X,Y) \leftarrow r_0(X, V_0)\\
    & S(p_{10}): r_t(sk_0,sk_1) \leftarrow r_1(sk_0, sk_2), r_0(sk_2, sk_3), r_0(sk_3, sk_4)
\end{align*}
we want to know whether $p_9 \preceq_{OI} p_{10}$. We first apply $\theta = \{X \backslash sk_0, Y \backslash sk_1 \}$ to $p_9$ such that $p_9[0]$ is eliminated because $p_9[0]\theta=S(p_{10})[0]$. With $p_9[1]\theta=r_0(sk_0, V_0)$, the proof procedure first tests $p_9[1]\theta$ against $S(p_{10})[2]$ because they share the same predicate $r_0$, which fails because $sk_0$ can not be reduced to $sk_2$. Then, the proof procedure backtracks to test $p_9[1]\theta$ against $S(p_{10})[3]$, which also fails. Therefore, we conclude $p_9 \not \preceq_{IO} p_{10}$.
\end{exp1}

The backtracking process introduces unnecessary complexities to the resolution of OI-subsumption among straight and connected rules. We propose a constrained version of OI-subsumption that simplifies the proof procedure by replacing the backtracking process with a position constraint. Also, it is sound and complete w.r.t OI-subsumption.

\begin{def1}[SA-Subsumption]
Sequence-aware subsumption (SA-Subsumption) is defined as: given rules $p$ and $p'$, we consider $p$ SA-subsumes $p'$, denoted by $p \preceq_{SA} p'$, iff:
\begin{align}
    & \forall i \in [0,|p|] : p[i]\theta = p'[i]
\end{align}
where notation $|\cdot|$ indicates the number of atoms in a rule, and variables substituted in $\theta$ refer to different entities.
\end{def1}

In comparison to the proof procedure of OI-subsumption that backtracks when the elimination test fails, SA-subsumption only needs to check if atoms at the same positions in rules $p$ and $p'$ are the same after substitution to decide the subsumption relationship between $p$ and $p'$.

\begin{prop1} \label{prop2.1}
On connected and straight rules, SA-subsumption is sound and complete w.r.t OI-subsumption.
\begin{proof}
Consider we have connected and straight rules $p$ and $p'$ where $|p| \le |p'|$, we want to know whether $p \preceq_{OI} p'$. We eliminate $p[i-1]=r_0(V_0, V_1)$ by matching $S(p')[i-1]=r_0(sk_0, sk_1)$ with $\theta=\{V_0 \backslash sk_0, V_1 \backslash sk_1 \}$ where the next atom $p[i]=r_0(V_1, V_2)$ is instantiated into $p[i]\theta=r_0(sk_1, V_2)$. Due to the straightness of rules, $sk_1$ can only occur at most once in the rest of $S(p')$ except for $S(p')[i-1]$, and according to the connectedness of rules, $sk_1$ can only occur in $S(p')[i]$. To eliminate $p[i]$, we need $p[i]\theta' = S(p')[i]$ or simply $p[i]\theta' = p'[i]$ where $\theta'$ is extended from $\theta$. Therefore, we prove when $p \preceq_{OI} p'$, $p \preceq_{SA} p'$. One special case is that, with rules $p_9$ and:
\begin{align*}
	& p_{11}: r_t(X, Y) \leftarrow r_0(Y, V_0), ..., r_0(X, V_n) \quad p_{12}: r_t(X, Y) \leftarrow r_0(X, V_n), ..., r_0(Y, V_0)
\end{align*}
we can prove $p_{9} \preceq_{OI} p_{11}$ yet $p_{9} \not \preceq_{SA} p_{11}$. We preserve the completeness by reversing the order of the body atoms in $p_{11}$ to create an equivalent rule $p_{12}$ that have $p_{9} \preceq_{OI} p_{12}$ and $p_{9} \preceq_{SA} p_{12}$. It is obvious that SA-subsumption is sound w.r.t OI-subsumption as when $p \preceq_{SA} p'$, we have $\exists \theta: p \theta \subseteq p'$, thus $p \preceq_{OI} p'$.
\end{proof}
\end{prop1}

\subsection{Proper Rule Hierarchy}

\begin{figure}[t]
\floatconts{fig.2}{\caption{An incomplete proper rule hierarchy generated based on the knowledge graph in Figure.\ref{fig.1a}. We use abbreviations for lengthy predicate names, where $A$ for "Advises", $I$ for "Is\_a" and $P$ for "Publishes". Dashed lines represent I-subsumption relationships and solid lines represent A-subsumption relationships.}}{
    \centering
    \includegraphics[width=1\textwidth]{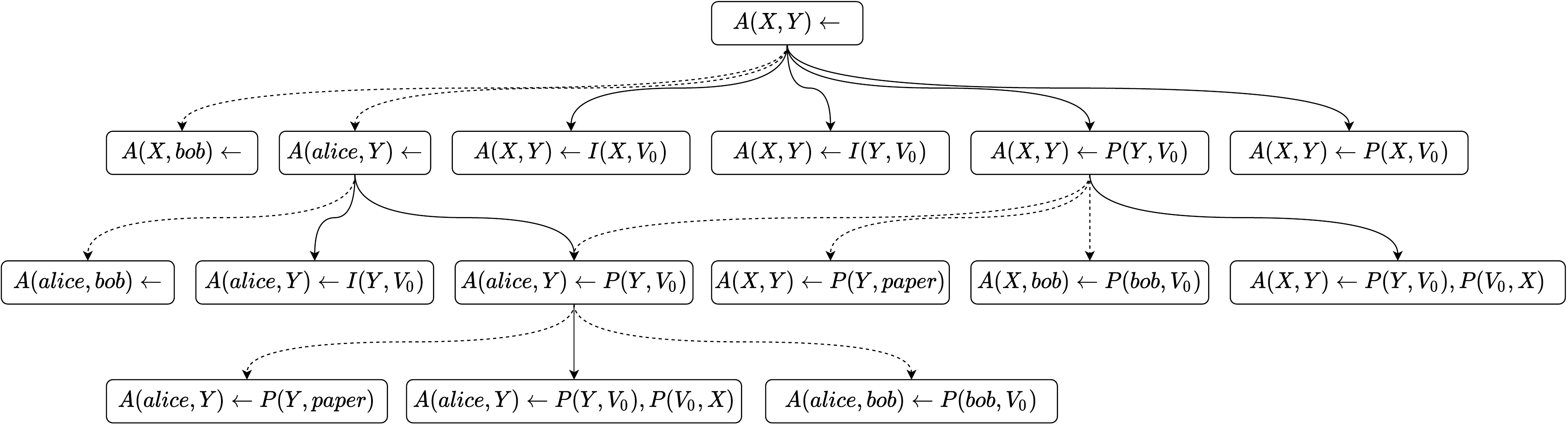}}
    % \label{fig.2}
\end{figure}

According to Proposition \ref{prop2.1} and Example \ref{exp2.3}, we can infer that $(\{p_4, p_7, p_8\}, \preceq_{SA})$ is also $\{(p_8, p_4), (p_7, p_8), (p_7, p_4)\}$. As SA-subsumption is transitive, knowing $(p_7,p_8)$ and $(p_8,p_4)$, relationship $(p_7,p_4)$ can be inferred via transitivity. We call a subsumption relationship \textit{redundant} if it can be inferred via transitivity, and a rule hierarchy \textit{proper} if it contains no redundant relationships. We here introduce an approach that builds a proper rule hierarchy $\Phi$ from a rule set $\mathcal{F}$ w.r.t SA-subsumption, that is $\forall (p, p') \in \Phi: p \preceq_{SA} p'$.

To build a proper rule hierarchy, we take inspirations from the top-down methods \citep{Galarraga2015} that simultaneously construct and explore rule hierarchies by repeatedly applying atomic specialization operators. An atomic specialization operator produces a rule $p'$ from another rule $p$ such that $p \preceq p'$ by modifying at most one element in $p'$ at a time. By employing the subsumption frameworks that only identify the subsumption relationships corresponding to the relationships implied by atomic specialization operators, we can map a set of rules to a proper rule hierarchy. By assuming the continuity of rule sets, that is, for instance, if rule $p_4$ exists, its generalization $p_8$ also exists in the rule set, we here propose two subsumption frameworks that are derived from SA-subsumption and correspond to the atomic specialization operators that perform atom addition and variable instantiation, respectively.

\begin{def1}[A-Subsumption] \label{def-asub}
Addition subsumption (A-Subsumption) identifies the relationship between rules $p$ and $p'$ implied by a specialization operator which adds a new atom that shares a connecting variable with the last atom in $p$ to $p$ to create $p'$. $p$ A-subsumes $p'$, denoted by $p \preceq_{A} p'$, iff $p \preceq_{SA} p'$, $d(p) = d(p')$ and $|p'|=|p|+1$.
\end{def1}

\begin{def1}[I-Subsumption] \label{def-isub}
Instantiation subsumption (I-Subsumption) identifies the relationship between rules $p$ and $p'$ implied by a specialization operator which instantiates a variable in $p$ to create $p'$. $p$ I-subsumes $p'$, denoted by $p \preceq_{I} p'$, iff $p \preceq_{SA} p'$, $d(p) = d(p') + 1$ and $|p|=|p'|$.
\end{def1}

Given a rule set $\mathcal{F}$, as the individual hierarchies $\Phi_a=(\mathcal{F}, \preceq_{A})$ and $\Phi_i=(\mathcal{F}, \preceq_{I})$ are proper and resolved by the constrained versions of SA-subsumption, combined hierarchy $\Phi=\Phi_{a} \cup \Phi_{i}$ is proper w.r.t SA-subsumption.

\begin{exp1}
Illustrated in Figure.\ref{fig.2} is the union of hierarchies ordered by A-Subsumption and I-Subsumption. It is a proper hierarchy w.r.t SA-subsumption. For instance, although the subsumption relationship $A(X,Y) \leftarrow \preceq_{SA} A(X,bob) \leftarrow P(bob, V_0)$ is valid, it is not explicitly linked in the hierarchy as it can be inferred via transitivity by considering $A(X,Y) \leftarrow \preceq_{SA} A(X,Y) \leftarrow P(Y, V_0)$ and $A(X,Y) \leftarrow P(Y, V_0) \preceq_{SA} A(X,bob) \leftarrow P(bob, V_0)$.
\end{exp1}

\section{Framework Adaptation} \label{sec.3}

In this section, we adapt RHF to GPFL \citep{Gu2020} which is a walk-based method optimized in mining instantiated rules, and design and implement two HPMs, namely prior pruning and post pruning, that utilize generated rule hierarchies to remove irrelevant and redundant rules, respectively. 

\subsection{GPFL Introduction}
GPFL is a discriminative walk-based rule learner that mines rules for one target predicate at a time. In contrast to other walk-based methods \citep{Lao2015,Ortona2018,Meilicke2019} that generate rules from paths by solely applying generalization, GPFL first generalizes paths into abstract rules, including CARs and OARs, and then specializes the OARs into two sub-types of INSRs:
\begin{align*}
    & \text{\textbf{HAR}}:\quad r_t(X, e_i) \leftarrow r_0(X, V_1), ..., r_n(V_n, V_{n+1})\\
    & \text{\textbf{BAR}}:\quad r_t(X, e_i) \leftarrow r_0(X, V_1), ..., r_n(V_n, e_j)
\end{align*}
A Head Anchored Rule (HAR) is a specialization of an OAR where the non-connecting variable in the head atom is substituted with a constant, and a Both Anchored Rule (BAR) is a specialization of a HAR where the non-connecting variable in the last body atom is replaced by a constant. The two-stage generalization-specialization rule generation mechanism allows GPFL to efficiently create and evaluate INSRs where structurally similar INSRs are collectively evaluated over shared groundings. It also presents an opportunity, by adopting RHF, to perform stage-wise hierarchical pruning to effectively remove unpromising rules for improved efficiency.

\begin{algorithm2e}[tb] 
\SetKwInOut{Input}{Input}\SetKwInOut{Output}{Output}
\SetKwRepeat{Do}{do}{while}
\SetKwFunction{Generalization}{Generalization}
\SetKwFunction{Specialization}{Specialization}
\SetKwFunction{PriorPruning}{PriorPruning}
\SetKwFunction{I-Subsumption}{I-Subsumption}
\SetKwFunction{PostPruning}{PostPruning}
\DontPrintSemicolon
\LinesNumbered

\Input{$\mathcal{G}, I, len$}
\Output{Mined rule set $\mathcal{F}$}

Initialize empty set $\mathcal{F}$\;
$L \leftarrow$ Generalization($\mathcal{G}, I, len$)\;
$\Phi_{a} \leftarrow$ A-Subsumption($L$)\;
$L' \leftarrow$ PriorPruning($\Phi_{a}$)\;
\For{$l \in L'$}{
    \If{$l$ is a CAR}{
        Check quality of $l$ and add $l$ to $\mathcal{F}$ if it is relevant\;
    }
    \Else{
    $S \leftarrow$ Specialization($l, \mathcal{G}, I$)\;
    Filter out irrelevant rules in $S$\;
    \If{$S$ is not empty after filtering}{
        $\Phi_{i} \leftarrow$ I-Subsumption($S$)\;
        $S' \leftarrow$ PostPruning($\Phi_{i}$)\;
        Add all rules in $S'$ to $\mathcal{F}$\;
    }
    }
}
\textbf{Return} $\mathcal{F}$\;
\caption{Augmented GPFL with Prior and Post Pruning}
\label{alg.1}
\end{algorithm2e}

\subsection{Algorithm}
In Algorithm.\ref{alg.1}, we introduce an augmented GPFL where we closely align the construction of rule hierarchies and the application of HPMs to the two-stage rule generation mechanism employed in GPFL. The system takes as inputs a knowledge graph $\mathcal{G}$, a set of positive instances of a target predicate $I$, and the maximum length a rule can have $len$ and outputs a rule set $\mathcal{F}$. The length of a rule is the number of its body atoms. In the procedure \texttt{Generalization($\mathcal{G}, I, len$)}, the system generalizes a set of paths originated from $I$ within length $len$ into OARs and CARs and adds them to a rule set $L$. It then applies A-subsumption to $L$ to build a rule hierarchy $\Phi_{a}$ that only contains the subsumption relationships between abstract rules. 

Before proceeding to discuss the prior pruning, we define a set of rule quality measures that characterize the relevancy of rules. Consider we have a rule $p$, the support \citep{Galarraga2015} of $p$ is defined as:
\begin{align}
    & supp(p) = |\{(x,y)|(x,y) \in g(p) \cap R_t \}|
\end{align}
where $g(p)$ is the head groundings inferred by grounding the body atoms of $p$ over $\mathcal{G}$, and $R_t$ is a set of ground atoms in $\mathcal{G}$ that are the instances of the target predicate $r_t$. We define the head coverage \citep{Galarraga2015} of $p$ as:
\begin{align}
    & hc(p) = \frac{supp(p)}{|R_t|}
\end{align}
and smooth confidence \citep{Meilicke2019} as:
\begin{align}
    & sc(p) = \frac{supp(p)}{\eta + |g(p)|}
\end{align}
where $\eta$ is an user-defined offset. We call a rule $p$ \textit{relevant} iff $supp(p) > supp_f$, $hc(p) > hc_f$ and $sc(p) > sc_f$ where $supp_f$, $hc_f$ and $sc_f$ are pre-defined thresholds for corresponding measures. We aim to use the procedure \texttt{PriorPruning($\Phi_a$)} to remove irrelevant INSRs before their creation. More specifically, as INSRs are generated by specializing the OARs, the system can avoid the creation and evaluation of irrelevant INSRs by identifying and pruning the OARs that potentially create irrelevant INSRs. Therefore, the problem is reduced to the efficient identification of unpromising OARs. In the \texttt{PriorPruning($\Phi_a$)}, the system traverses the hierarchy $\Phi_a$ starting from the top rule, which is the rule that does not have any ancestors, in a breadth-first fashion. A visited rule $p$ and all of its descendants are pruned if $supp(p) < supp_h$ where the prior threshold $supp_h$ is a pre-defined threshold on support. This makes sense because the support of rules is a monotonic measure that tends to become smaller with increasing depth in a rule hierarchy. 

After the application of prior pruning, the system iterates over rules in the filtered rule set $L'$. For a rule $l$ in $L'$, if it is a CAR and considered relevant after evaluation, it will be added to the rule set $\mathcal{F}$. If $l$ is an OAR, the procedure \texttt{Specialization($l, \mathcal{G}, I$)} is applied where a set of HARs and BARs $S$ are derived from $l$ by instantiating certain variables in $l$. Irrelevant rules in $S$ are then filtered out. We call an OAR \textit{informative} if its $S$ is not empty after the removal of irrelevant rules, otherwise it is \textit{uninformative}. Inspired by the hierarchical feature selection approaches proposed in \citet{Ristoski2014}, we utilize a simple mechanism to identify and prune redundant rules. Given a rule hierarchy $(\mathcal{F}, \preceq)$ over $\mathcal{G}$, we consider a rule $p$ \textit{redundant} if:
\begin{align}
    & \exists p' \in \mathcal{F}: p' \preceq p, sc(p') > sc(p)
\end{align}
We argue that as $g(p) \subseteq g(p')$ because of $p' \preceq p$, and $sc(p') > sc(p)$, the existence of $p$ does not provide new information to the reasoning over $\mathcal{G}$. We corroborate the argument via the experiment results in the following section. By applying I-subsumption to $S$, the generated hierarchy $\Phi_{i}$ contains pairs of HARs and BARs that share subsumption relationships. In the procedure \texttt{PostPruning($\Phi_{i}$)}, by removing the BARs that have smaller confidence than the HARs subsuming them, the system creates a set $S'$ free of redundant rules, and then adds all rules in $S'$ to $\mathcal{F}$. Eventually, $\mathcal{F}$ is returned as the learned rule set where irrelevant and redundant rules are removed by the prior and post pruning, respectively.

\section{Experiments}

In this section, we use the augmented GPFL as an example to demonstrate the effectiveness of the application of HPMs through experiments over four publicly available datasets in two settings. By enabling the application of the prior and post pruning, we observe considerable reductions in the runtime and the number of learned rules without compromising the predictive performance.

\begin{table}[t]
\floatconts{tab.1}{\caption{Statistics of the benchmark datasets.}}{
\centering
\scalebox{0.8}{
\begin{tabular}{lrrrrrr}
\toprule
& & & \multicolumn{4}{c}{\#Triples}\\
\cmidrule{4-7}
Dataset & \#Entities & \#Types & \#Train & \#Valid & \#Test & \#Total\\
\midrule
FB15K-237-LV & 14.54K & 237 & 185K & 62.02K & 62.12K & 310K\\ 
NELL995-LV & 75.49K & 200 & 92.44K & 30.84K & 30.92K & 154K\\
WN18RR-LV & 40.94K & 11 & 55.79K & 18.60K & 18.60K & 93K\\
OBL-PE & 180K & 28 & 4.19M & 183K & 180K & 4.55M\\
\bottomrule
\end{tabular}
}
}
\end{table}

\subsection{Datasets}

We select four publicly available datasets for experiments, including FB15K-237 \citep{Toutanova2015}, WN18RR \citep{Dettmers2018}, NELL995 \citep{Xiong2017} and OBL-PE \citep{Breit2020}. FB15K-237, WN18RR and NELL995 are popular datasets for evaluating the performance of methods on Knowledge Graph Completion (KGC) task, and OBL-PE is a subset of OBL dataset \citep{Breit2020} that only contains positive triples. FB15K-237, NELL995 and OBL-PE are created in such a way that their validation and test sets contain no triples that are the reverse of known triples in the training set. These reverse triples allow models with trivial rule $r_t(X,Y) \leftarrow r_t(Y,X)$ to perform exceptionally well, which makes it hard to understand the true performance of different approaches. WN18RR has around 35\% of the triples in validation and test sets that are reverse triples.

As pointed out in \citet{Gu2020}, the performance of instantiated rule learners suffers greatly from the appearance of overfitting rules, and to remove overfitting rules through validation, large validation set is needed. However, the sizes of validation sets in the default splits of FB15K-237, NELL995 and WN18RR are too small to be useful for validation. Specifically, the validation triples to total triples ratio is 5\% for FB15K-237, 0.3\% for NELL995 and 3\% for WN18RR. In this work, we re-split FB15K-237, NELL995 and WN18RR into training/validation/test sets in a 6:2:2 ratio, and rename them by adding a "-LV" suffix that stands for large validation. Based on our observations, the performance of the re-split OBL-PE is similar to that of the original splits, so we keep the original OBL-PE for experiments. After re-splitting, the proportion of reverse triples is 6.5\% for FB15K-237-LV, 19\% for WN18RR-LV, and 6\% for NELL995-LV. Statistics about these datasets is listed in Table.\ref{tab.1}.

\begin{figure}
\centering
\floatconts{fig.3}{\caption{Diagrams that show the numbers of different types of OARs in the sets of learned rules over experiment datasets, where P-OARs stands for pruned OARs; I-OARs for informative OARs, and U-OARs for uninformative OARs. The prior threshold is set to 10.}}{\includegraphics[width=0.95\textwidth]{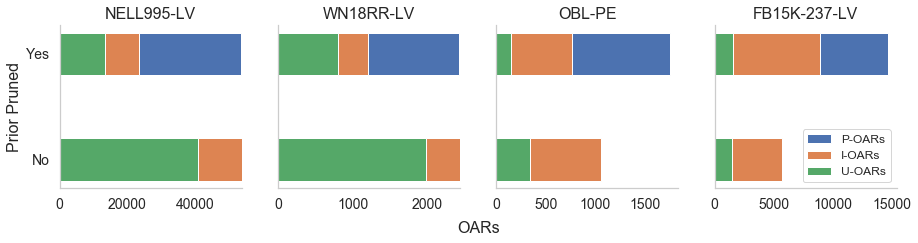}}
\end{figure}

\subsection{Experiment Setup}

We implement RHF and the HPMs on top of the GPFL codebase\footnote{\url{https://github.com/irokin/GPFL}}. GPFL is implemented in Java and deeply integrated with the Neo4j\footnote{\url{https://github.com/neo4j/neo4j}} graph database. We configure GPFL to run in both constrained and unconstrained settings. For the majority of rule learners, it is often not possible to explore the entire rule space in a reasonable time on large KGs. Therefore, various time and space constraints are adopted to terminate systems prematurely to meet specific time and space requirements. For experiments on FB15K-237-LV and OBL-PF, we put time constraints on the generalization and specialization procedures, that is when the time constraints are reached, the system stops the running procedure and proceeds. For WN18RR-LV and NELL995-LV, we configure the system to run without constraints. For all experiments, we use following setting of parameters: $len$ for CARs and INSRs is set to 3, and the filtering of overfitting rules is turned on where the overfitting threshold is set to 0.1. Except for WN18RR-LV, we set $supp_f$ to 3, $hc_f$ to 0.001 and $sc_f$ to 0.001. For WN18RR-LV, we set $supp_f$ to 2, $hc_f$ to 0.0001 and $sc_f$ to 0.0001. All experiments are conducted on the AWS EC2 instances that have 8 CPU cores and 64GM RAM. We have made our codebase and datasets available at \url{https://github.com/irokin/RuleHierarchy}.

\begin{figure}[t]
\centering
\floatconts{fig:example2}% 
{\caption{Experiment results over different prior thresholds on the augmented GPFL. When the threshold is 0, it reports the results of the original GPFL.}}% 
{%
\subfigure[NELL995-LV][centred]{%
\label{fig4a}% label for this sub-figure
\includegraphics[scale=0.35]{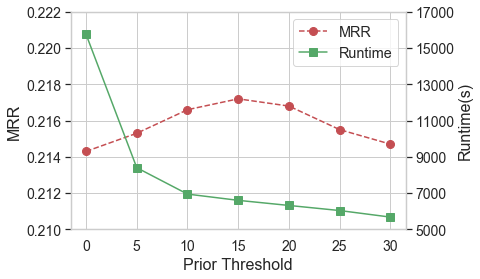}
}\quad % space out the images a bit
\subfigure[WN18RR-LV][centred]{%
\label{fig4b}% label for this sub-figure
\includegraphics[scale=0.35]{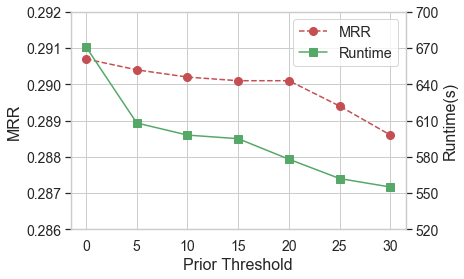}
}
\subfigure[FB15K-237-LV][centred]{%
\label{fig4c}% label for this sub-figure
\includegraphics[scale=0.35]{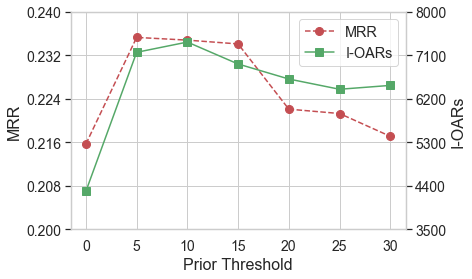}
}\quad % space out the images a bit
\subfigure[OBL-PE][centred]{%
\label{fig4d}% label for this sub-figure
\includegraphics[scale=0.35]{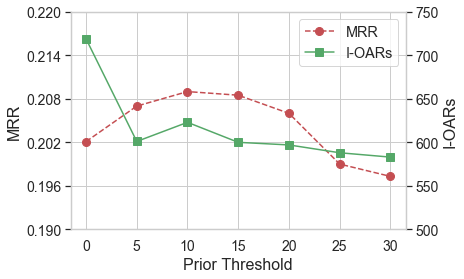}
}
}
\label{fig.4}
\end{figure}

\subsection{Evaluation Protocol}

We evaluate the predictive performance of the system by tasking it with the KGC problem. Specifically, A KGC query takes the form of $r_t(e_i, ?)$ or $r_t(?, e_i)$ where $r_t$ is the target predicate, and the question mark is expected to be replaced with the candidates $e \in \mathcal{E}$ that are suggested by learned rules such that predictions $r_t(e_i, e)$ or $r_t(e, e_i)$ for $r_t$ are proposed. We follow the evaluation protocol proposed in \citet{Bordes2013} where both head query $r_t(e_i, ?)$ and tail query $r_t(?, e_i)$ are answered. For ranking the predictions of a query, we adopt the maximum aggregation strategy proposed in \citet{Meilicke2019} where predictions are sorted by the maximum of the confidence of rules suggesting the predictions, if there are ties, the tied predictions are resolved by recursively comparing the next highest confidence of suggesting rules until all ties are resolved. We report experiment results in Mean Reciprocal Rank (MRR) in filtered setting \citep{Bordes2013}. For FB15K-237-LV and NELL995-LV, we randomly select 20 target predicates for experiments, and for WN18RR-LV and OBL-PE, all predicates are used for experiments.

\subsection{Prior Pruning}

In this section, we show that the prior pruning is able to effectively remove irrelevant rules. All experiments are conducted without the application of the post pruning where when the prior threshold is set to 0 or turned off, the augmented GPFL is reduced to the original GPFL which serves as the baseline.

\begin{table}[t]
\floatconts{tab.2}{\caption{Post pruning experiment results. \#R-Rules is the number of relevant rules and RAT stands for the rule application time measured in seconds.}}{
    \centering
    \scalebox{0.85}{
    \begin{tabular}{lrrrr}
    \toprule
    Dataset & PostPrune & MRR & \#R-Rules  & RAT(s)\\
    \midrule
    \multirow{2}{*}{FB15K-237-LV} & No & 0.215 & 6.07M  & 4.17K\\
    & Yes & 0.216 & 5.11M  & 3.56K\\
    \midrule
    \multirow{2}{*}{OBL-PE} & No & 0.202 & 8.69M & 2.75K\\
    & Yes & 0.202 & 7.63M & 2.67K\\
    \midrule
    \multirow{2}{*}{NELL995-LV} & No & 0.214 & 5.59M & 4.78K\\
    & Yes & 0.214 & 4.08M & 4.27K\\
    \midrule
    \multirow{2}{*}{WN18RR-LV} & No & 0.291 & 28.9K & 39.5\\
    & Yes & 0.291 & 28.8K & 38.7\\
    \bottomrule
    \end{tabular}
    }
    }
\end{table}

As prior pruning removes irrelevant rules by identifying and pruning unpromising OARs, and uninformative OARs (U-OARs) are the most unpromising OARs in that none of the rules derived from U-OARs are relevant, we use the number of U-OARs as an indicator to the effectiveness of the prior pruning. We start by discussing the experiment results on NELL995-LV and WN18RR-LV that are conducted in the unconstrained setting. As illustrated in Figure.\ref{fig.3}, on both NELL995-LV and WN18RR-LV, the numbers of U-OARs drop significantly while the numbers of informative OARs (I-OARs) stay more or less the same. It is clear that U-OARs account for most of the pruned OARs (P-OARs). Accordingly, as shown in Figure.\ref{fig4a} and Figure.\ref{fig4b}, the removal of U-OARs by the prior pruning results in significant decrease in the runtime where the predictive performance fluctuates within a small range. In the constrained setting, the effect of the removal of irrelevant rules is more complicated due to the interactions among various factors. In Figure.\ref{fig.3}, on both OBL-PE and FB15K-237-LV, the total number of visited OARs with the prior pruning applied is considerably larger than that of the baseline because the runtime wasted on creating and evaluating irrelevant rules is allocated to the discovery of more rules. In particular on FB15K-237-LV, the saved runtime leads to a 72\% growth in the number of I-OARs, whereas on OBL-PE, most of the newly discovered rules are U-OARs. In Figure.\ref{fig4c} and Figure.\ref{fig4d}, as in constrained setting the runtime of the experiments with different thresholds is always the same, we instead compare the number of I-OARs. Specifically, we observe on FB15K-237-LV that the growth in I-OARs contribute to the improvements in the predictive performance, and on OBL-PE, because most of the I-OARs are already discovered with the baseline, its performance gain is negligible. It is worth noting that the prior pruning is much more effective on the datasets with large amount of relationship types. This is reasonable because the prior pruning operates in the space of abstract rules and the amount of abstract rules is proportional to that of relationship types. 

\subsection{Post Pruning}

In contrast to the prior pruning that aims to remove irrelevant rules, the post pruning aims to remove relevant rules that are considered redundant. We demonstrate the validation of our definition of redundant rules and the effectiveness of the post pruning by examining if the predictive performance is affected by the removal of relevant rules. As demonstrated in Table.\ref{tab.2}, the removal of relevant rules has negligible impact on the predictive performance while the rule application time (RAT) is reduced for improved efficiency. It is worth noting that on FB15K-237-LV, 15\% of the relevant rules are removed by the post pruning at the expense of a decrease by 0.01 in MRR, which consequently reduces the RAT by around 600s. Post pruning is much less effective on WN18RR-LV because most of its relevant rules have larger confidence than their ancestors.

\section{Conclusion}

In this work, we aim to apply HPMs to walk-based rule learners for improved efficiency. To achieve this, we first introduce RHF that constructs a proper rule hierarchy from a set of rules using a collection of subsumption frameworks, and then adapt RHF to GPFL where we design and implement two HPMs that utilize the generated rule hierarchies to remove unpromising rules. Through experiments over four datasets, we use the augmented GPFL to demonstrate the effectiveness of the application of HPMs, where we observe significant reductions in the runtime and the number of learned rules without compromising the predictive performance. This successful adaptation demonstrates the benefits and practicability of applying HPMs to walk-based methods. By taking this work as a foundation, developing more HPMs for walk-based methods is a possible future research direction.  

\bibliography{bibfile}

\end{document}